\documentclass{article}

\PassOptionsToPackage{numbers,compress}{natbib}
\usepackage[preprint]{neurips_2026}

\usepackage[utf8]{inputenc}
\usepackage[T1]{fontenc}
\usepackage{hyperref}
\usepackage{url}
\usepackage{booktabs}
\usepackage{amsfonts}
\usepackage{nicefrac}
\usepackage{microtype}
\usepackage{xcolor}
\usepackage{amsmath,amssymb}
\usepackage{graphicx}
\usepackage{algorithm}
\usepackage{algorithmic}
\usepackage{multirow}
\usepackage{xspace}
\usepackage{here}
\usepackage{subcaption}
\usepackage{caption}
\usepackage{placeins}

\title{Full-Stack FP4: Stable LLM Pretraining with Quantized Projections, Optimizers, and Attention}

\author{%
  Siyu Ding$^1$ \\
  \And
  Mingchuan Ma$^2$ \\
  \And
  Jiabo Tong$^{1,3}$ \\
  \AND
  Xingrun Xing$^1$ \\
  \And
  Ziming Wang$^4$ \\
  \And
  Guoqi Li$^{1,5,6}$ \thanks{Corresponding author: \texttt{guoqi.li@ia.ac.cn}} \\
}
\date{}

\begin{document}

\maketitle

\hspace{10pt}
\begin{minipage}{0.75\linewidth}
\small
\rmfamily
$^1$Institute of Automation, Chinese Academy of Sciences\\
$^2$Sichuan University\\
$^3$Beijing Zhongguancun Academy\\
$^4$Zhejiang University\\
$^5$Beijing Key Laboratory of Brain-Inspired General Intelligence Large Model\\
$^6$Key Laboratory of Brain Cognition and Brain-inspired Intelligence Technology
\end{minipage}
\vspace{10pt}

\begin{abstract}
\setcounter{footnote}{0}
Recent NVFP4 pretraining work has primarily optimized Transformer linear projections, leaving persistent optimizer states, optimizer computation, and low-precision attention forward--backward paths less explored. We present \textbf{Full-Stack FP4}, a modular NVFP4 framework with separate recipes for projections, AdamW states, Root/Muon computation, and attention. \textbf{LoRA-SVD} protects a compact projection subspace in BF16 while retaining full-shape NVFP4 computation, reducing the linear-only loss gap from \textbf{1.40\%} to \textbf{0.61\%}. An ordered square-root, tile-mean, and Hadamard pipeline enables stable NVFP4 AdamW momentum storage; shape-dependent coefficients and clipping stabilize direct NVFP4 Root iterations; and mixed-precision attention retains softmax-sensitive operations in BF16. On 3B pretraining with 64B tokens, BF16 and Full-Stack FP4 reach losses of \textbf{2.267} and \textbf{2.286}, a \textbf{0.838\%} gap. Their average zero-shot perplexities are 26.675 and 26.665, respectively, with Full-Stack FP4 averaging 0.10 percentage points lower in accuracy. Native four-block measurements on one RTX 5090 show 2.50--2.83$\times$ Root speedups over optimized BF16 and 37.9--42.5\% lower AdamW peak memory.
\end{abstract}

\section{Introduction}
\label{sec:intro}

The computational cost of LLM pretraining has driven precision down from FP32 to BF16 and FP8, yielding substantial memory and throughput gains. Blackwell GPUs further provide native NVFP4 Tensor Core support and make 4-bit pretraining a practical systems target. Recent work has demonstrated stable 4-bit pretraining for Transformer linear layers~\citep{nvfp4,fp4alltheway2025,quartet2025,chon2026,fouroversix2025,tetrajet2025,chen2025tetrajetv2}, and low-bit Muon updates have also begun to receive attention~\citep{wu2026achieving}. However, most NVFP4 recipes retain higher-precision optimizer states, optimizer computation, and attention.

The main difficulty is that these modules do not share one numerical structure. Projection error remains relevant as optimization signals weaken later in training: FP4-All-the-Way~\citep{fp4alltheway2025} observes that gradient signals decrease while quantization noise remains approximately constant, reducing the effectiveness of continued NVFP4 training. AdamW's second moment is non-negative, heavy-tailed, persistent, and denominator-sensitive. Root applies repeated Newton--Schulz matrix products, whereas attention contains softmax-sensitive paths and must preserve forward--backward tensor consistency. A uniform quantization rule is therefore insufficient.

We present Full-Stack FP4, a collection of independent and composable low-precision recipes for these existing training modules. LoRA-SVD protects a compact principal subspace in BF16 while leaving the dominant dense computation in NVFP4. By systematically reducing projection error at low overhead, it may extend the useful NVFP4 training phase, although our experiments do not establish long-horizon behavior. Quantized-state AdamW reshapes its second moment before NVFP4 storage. NVFP4 Root uses shape-dependent coefficients and outlier clipping to stabilize direct Newton--Schulz iterations. Mixed-precision attention quantizes suitable $Q/K$ and backward paths while retaining sensitive operations in BF16; Appendix~\ref{app:method_detail:sensitivity_analysis} analyzes the $PV$ and $dOV^\top$ paths.

Our technical contributions are:
(1) \textbf{LoRA-SVD projections}, a mergeable training-time reparameterization that reduces the linear-only loss gap from \textbf{1.40\%} to \textbf{0.61\%};
(2) \textbf{quantized-state AdamW}, an optimizer-state-specific NVFP4 pipeline supported by component-level reconstruction analysis and native-training stability;
(3) \textbf{NVFP4 Root}, a direct low-precision Newton--Schulz path enabled by shape-dependent coefficients and clipping;
(4) \textbf{trainable mixed-precision attention}, which protects softmax-sensitive paths while quantizing suitable forward and backward computation.
We evaluate their joint behavior at 3B over 64B tokens and measure native four-block efficiency on an RTX 5090 against optimized BF16 and NVIDIA Transformer Engine (TE). The recipes can be enabled independently or combined; ``Full-Stack'' denotes their aggregate precision coverage rather than a monolithic deployment requirement.

Table~\ref{tab:comparison} summarizes the scope of representative work. Most NVFP4 pretraining methods focus on linear projections and retain higher-precision optimizer states and attention~\citep{nvfp4,fp4alltheway2025,quartet2025,chon2026,fouroversix2025,tetrajet2025,chen2025tetrajetv2}. COAT~\citep{coat2024} stores AdamW states in FP8, while low-bit Muon work~\citep{wu2026achieving} studies optimizer computation. Existing FP4 attention systems primarily target inference rather than a consistent NVFP4 forward--backward pretraining operator. Our work evaluates several missing module combinations and their joint convergence.

\begin{table}[t]
\centering
\caption{\textbf{Training-module coverage of representative low-precision methods.}
Wt/Act/G denotes projection weights, activations, and gradients; Train. Attn denotes a low-precision attention forward--backward path. ``Pretrain Eval.'' indicates that the covered components are evaluated in pretraining. SageAttn3 uses INT8 for its training path and FP4 for inference. $\checkmark$ = addressed; --- = not addressed; $\circ$ = partial or higher-precision coverage.}
\label{tab:comparison}
\small
\begin{tabular}{lcccccc}
  \toprule
  \textbf{Method} & \textbf{Prec.} & \textbf{Wt/Act/G} & \textbf{Opt.} & \textbf{Opt.} & \textbf{Train.} & \textbf{Pretrain} \\
   & & & \textbf{State} & \textbf{Comp.} & \textbf{Attn} & \textbf{Eval.} \\
  \midrule
  COAT~\citep{coat2024} & FP8 & $\checkmark$ & $\checkmark$ & --- & --- & $\checkmark$ \\
  NVFP4~\citep{nvfp4} & FP4 & $\checkmark$ & --- & --- & --- & $\checkmark$ \\
  FP4-All-the-Way~\citep{fp4alltheway2025} & FP4 & $\checkmark$ & --- & --- & --- & $\checkmark$ \\
  Quartet~\citep{quartet2025} & FP4 & $\checkmark$ & --- & --- & --- & $\checkmark$ \\
  CHON~\citep{chon2026} & FP4 & $\checkmark$ & --- & --- & --- & $\checkmark$ \\
  4/6~\citep{fouroversix2025} & FP4 & $\checkmark$ & --- & --- & --- & $\checkmark$ \\
  TetraJet-v2~\citep{chen2025tetrajetv2} & FP4 & $\checkmark$ & --- & --- & --- & $\checkmark$ \\
  Metis~\citep{cao2025metis} & FP4 & $\checkmark$ & --- & --- & --- & $\checkmark$ \\
  Low-bit Muon~\citep{wu2026achieving} & FP4 & --- & $\checkmark$ & $\checkmark$ & --- & --- \\
  SageAttn3~\citep{sageattn3_2025} & INT8/FP4 & --- & --- & --- & $\circ$ & --- \\
  \midrule
  \textbf{Ours} & \textbf{FP4} & $\checkmark$ & $\checkmark$ & $\checkmark$ & $\checkmark$ & $\checkmark$ \\
  \bottomrule
\end{tabular}
\end{table}

\section{Preliminaries}
\label{sec:prelim}
This section establishes the notation and background for the main computation primitives in our recipe.

\textbf{NVFP4 format.}
NVFP4 format~\citep{nvfp4} uses E2M1 encoding (1 sign,
2 exponent, 1 mantissa) together with \emph{block-wise} scaling: each block of 16x16 or 1x16 elements shares an FP8~E4M3 scale factor $s$, and an additional FP32
global scale $s_g$:
$\hat{x} = x_\mathrm{E2M1} \times s \times s_g$.
This finer granularity yields lower quantization error~\citep{nvfp4} compared to MXFP4's block-32 with E8M0 scale~\citep{mxfp2023}.

\textbf{Linear projection.}
A transformer linear layer computes $Y = XW^\top$ where
$W$ is the weight and
$X$ is the input activation.
Training involves three GEMMs: forward,
activation-gradient ($\nabla_X = \nabla_Y W$), and
weight-gradient ($\nabla_W = \nabla_Y^\top X$).

\textbf{AdamW optimizer.}
AdamW maintains two state tensors per parameter: the first-order momentum $m_t = \beta_1 m_{t-1} + (1{-}\beta_1)g_t$ and the second-order momentum $v_t = \beta_2 v_{t-1} + (1{-}\beta_2)g_t^2$. The update rule 
requires storing both $m_t$ and $v_t$ throughout training, adding two persistent state tensors per parameter.

\textbf{Root optimizer.}
Root~\citep{root2025} improves Muon~\citep{kosson2024muon} with
soft-thresholding for first-order momentum outlier removal and a Newton--Schulz (NS) orthogonalization iteration, $X_{k+1} = a X_k + b X_k (X_k^\top X_k) + c X_k (X_k^\top X_k)^2$, with optimized coefficients $a$, $b$, and $c$.
Root/Muon is not applied to every parameter: embeddings, \texttt{lm\_head}, normalization weights, biases, LoRA factors, and narrow parameters use AdamW.
The dense matmuls per iteration (typically $5$ cycles) dominate Root's compute cost.

\textbf{Attention operator.}
Standard multi-head attention computes
$S = QK^\top/\sqrt{d}$, $P = \mathrm{softmax}(S)$, $O = PV$
for each head, where $Q, K, V \in \mathbb{R}^{N \times d}$.
The most commonly used operator, FlashAttention~\citep{dao2022flashattention,dao2023flashattention2}, implements attention in a tiled, I/O-aware manner that avoids materializing the full $N \times N$ matrix.
The backward pass recomputes $S = QK^\top/\sqrt{d}$ and $P$, and computes $dV = P^\top dO$, $dP = dO \cdot V^\top$, $dS = P \odot (dP - \mathrm{rowsum}(dO \odot O))$, $dQ = dS \cdot K / \sqrt{d}$, $dK = dS^\top \cdot Q / \sqrt{d}$.
Overall, there are 2 matmuls in the forward pass and 5 in the backward pass, and the activation tensors ($Q$, $K$, $V$) dominate the memory and I/O cost.

\textbf{Others.}
We use two rounding strategies. Round-to-nearest (RTN) deterministically maps values to the closest representable number but can introduce systematic bias. Stochastic rounding (SR) is unbiased at the cost of added randomness. Hadamard transforms redistribute localized outlier energy and admit low-cost butterfly implementations. RTN is typically used in the forward pass, while SR and Hadamard transforms are applied to backward gradients~\citep{nvfp4,tseng2025mxfp4,fouroversix2025}. FP4-All-the-Way~\citep{fp4alltheway2025} identifies a gradient-norm threshold $\|\mathbf{g}\| < \sqrt{3}\sigma_q$, below which the weakening optimization signal falls beneath approximately constant quantization noise and continued FP4 training becomes less effective.

\section{Method}
\label{sec:method}
\subsection{Recipe Overview}
\label{sec:method:overview}

\begin{center}
\begin{tabular}{ccc}
  \toprule
  \textbf{Module} & \textbf{Numerical difficulty} & \textbf{Recipe} \\
  \midrule
  Linear projection (\S\ref{sec:method:linear})
    & residual projection error
    & NVFP4 residual + BF16 LoRA-SVD \\
  AdamW state (\S\ref{sec:method:adamw})
    & persistent, heavy-tailed moments
    & transformed NVFP4 states \\
  Root (\S\ref{sec:method:muon})
    & iterative low-precision error
    & shape-aware NVFP4 NS \\
  Attention (\S\ref{sec:method:attn})
    & softmax-sensitive paths
    & selective NVFP4/BF16 \\
  \bottomrule
\end{tabular}
\end{center}

Each recipe modifies an existing pretraining module and can be enabled independently. LoRA-SVD replaces selected linear layers during training; Q-AdamW changes the representation and update of existing momentum states; Q-Root changes the numerical path for eligible matrix updates; and Q-Attn changes precision assignments within attention. ``Full-Stack'' denotes their combined coverage, not a tightly coupled system that must be deployed at once. All recipes build on standard primitives including RTN, SR, random Hadamard transforms, and adaptive 4/6 scaling.

\begin{figure}[t]
  \centering
  \includegraphics[width=0.9\linewidth]{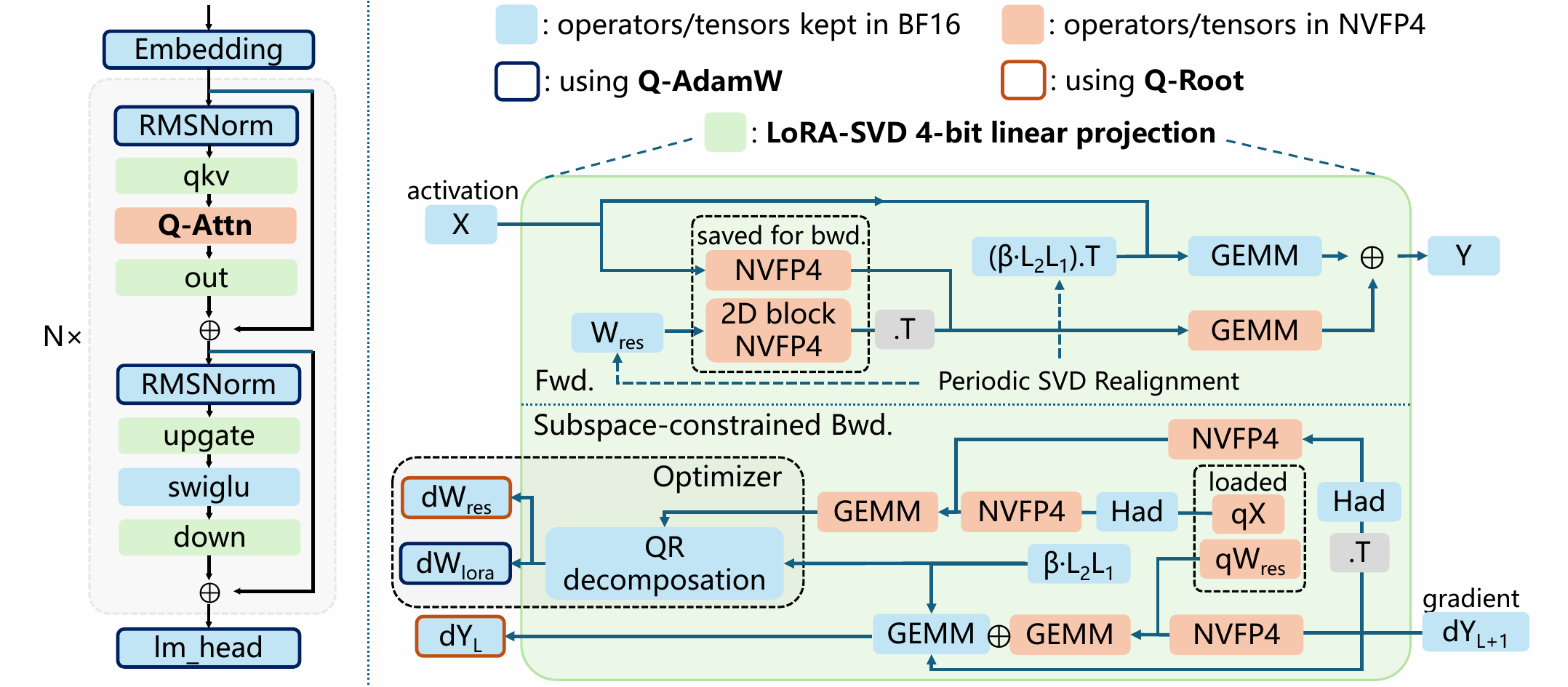}
  \caption{\textbf{Full-Stack FP4 overview.}
  LoRA-SVD protects a low-rank projection subspace; Q-AdamW compresses persistent momentum states; Q-Root executes eligible matrix updates through NVFP4 Newton--Schulz iterations; and Q-Attn assigns NVFP4 and BF16 to different attention paths. The recipes are independent and composable.}
  \label{fig:system_overview}
\end{figure}

\subsection{LoRA-SVD Linear Projections}
\label{sec:method:linear}

FP4-All-the-Way~\citep{fp4alltheway2025} observes that gradient signals weaken during training while quantization noise remains approximately constant, which can reduce the effectiveness of continued NVFP4 training. Reducing projection error without sacrificing the dominant low-precision computation may therefore extend the NVFP4 training phase. We do not evaluate this long-horizon hypothesis here.

LoRA-SVD parameterizes a training-time weight as
$W=W_{\mathrm{res}}+\beta L_2L_1$. It preserves a compact principal subspace and its activation path in BF16 while computing the full-shape residual in NVFP4. The parameterization directly replaces \texttt{nn.Linear}, does not alter the Transformer topology, and can be merged into a standard dense weight after training. The fusion strategy follows the low-rank traffic-reuse principle demonstrated by SVDQuant~\citep{li2024svdquant}; implementation details are discussed in Appendix~\ref{app:method_linear_kernel}.

\textbf{Subspace-constrained backward pass.}
To maintain alignment with the principal subspace, we use a Cholesky-QR constraint. The forward and backward computations are
\begin{gather}
Y = \underbrace{Q_\mathrm{1x16}(X)\,Q_\mathrm{16x16}(W_\mathrm{res})^\top}_{\text{NVFP4 GEMM}}
    + \,\beta \cdot \underbrace{(XL_1^\top)L_2^\top}_{\text{BF16}}
  \nonumber\\
  \mathrm{d} X = Q_{\mathrm{1x16}}(\mathrm{d} Y)\,\underbrace{Q_{\mathrm{16x16}}(W_{\mathrm{res}})}_{\text{reused from fwd.}} + \beta \cdot (\mathrm{d} Y\,L_2)\,L_1
  \label{eq:lorasvd-bwd-dx-naive} \\
  \mathrm{d} W = Q_{\mathrm{1x16}}(\mathrm{d} Y^\top H_{16}^r) \, Q_{\mathrm{16x1}}\left({H_{16}^r}^\top Q_{\mathrm{1x16}}(X)\right)
  \label{eq:lorasvd-bwd-dw-naive} \\
Q_{L_2} R_{L_2} = L_2, \quad Q_{L_1} R_{L_1} = L_1^\top \label{eq:qr} \\
\Lambda := Q_{L_2}^\top \mathrm{d}W Q_{L_1}, \qquad
\mathrm{d}W_{\mathrm{LoRA}} = Q_{L_2} \Lambda Q_{L_1}^\top \nonumber\\
\mathrm{d}W_{\mathrm{res}} = \mathrm{d}W - \mathrm{d}W_{\mathrm{LoRA}} \nonumber\\
\mathrm{d} L_1 = \beta L_2^\top \mathrm{d} W_{\mathrm{LoRA}},\quad
  \mathrm{d} L_2 = \beta \mathrm{d} W_{\mathrm{LoRA}} L_1^\top .
  \label{eq:lorasvd-bwd-dl-qr}
\end{gather}
$H_{16}^r$ is a fresh $16\times16$ random Hadamard transform. Components aligned with the maintained principal subspace update the BF16 low-rank branch, while the complementary residual updates $W_{\mathrm{res}}$. Equations~\ref{eq:qr}--\ref{eq:lorasvd-bwd-dl-qr} are deferred to the optimizer to avoid repeated work during gradient accumulation. Quantized activations are reused when forming LoRA gradients to avoid an additional high-precision activation read and large GEMM; Appendix~\ref{app:method_linear_detail} gives the derivation.

\begin{figure}[t]
  \centering
  \includegraphics[width=\linewidth]{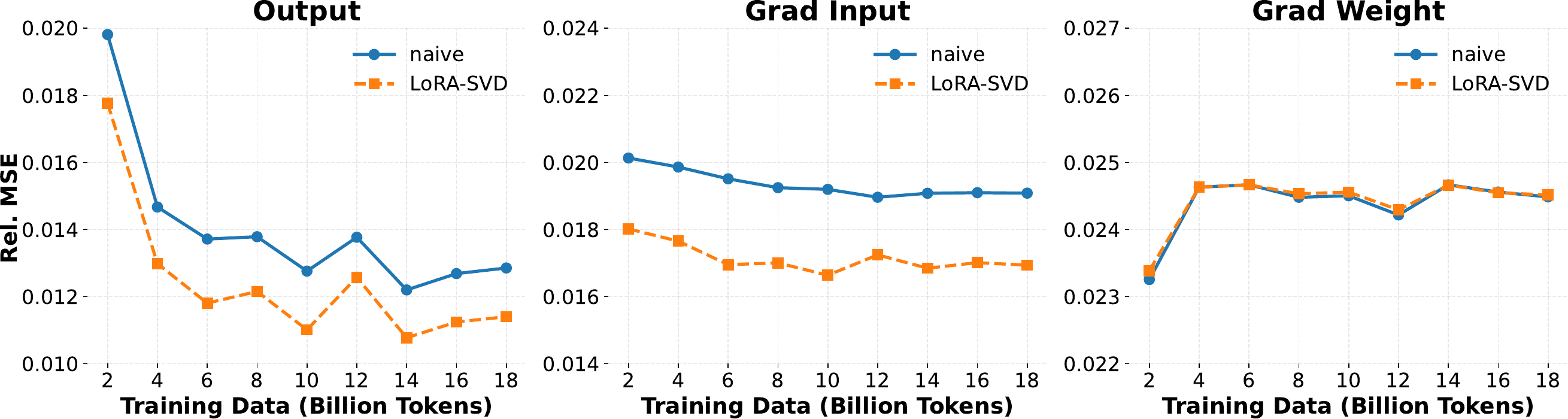}
  \caption{\textbf{Linear-projection error.}
  Each point averages relative MSE across sampled linear projections from the same 1B BF16 checkpoint. LoRA-SVD reduces error for outputs, input gradients, and weight gradients by approximately 10\% compared with direct NVFP4, with no increasing trend over the observed training trajectory.}
  \label{fig:linear_mse}
\end{figure}

Figure~\ref{fig:linear_mse} evaluates projection-level relative MSE using checkpoints from the 1B BF16 baseline. At each checkpoint, each plotted value averages relative MSE across the sampled linear projections for the corresponding output or gradient tensor. LoRA-SVD reduces the error of outputs, input gradients, and weight gradients throughout the observed training trajectory. In a separate paired-forward check on real text, quantized and unquantized forwards from the same checkpoint and input show 29.8--33.0\% lower final-output MSE with LoRA-SVD across $d=512$--$2048$.

\textbf{Periodic realignment.}
We realign the subspace every 2,048 steps using randomized SVD~\citep{halko2011finding} and reset $W_{\mathrm{res}},L_2,L_1$ using the top-$r$ components. At $d=4096/6144$, a trigger costs 160.00/226.96 ms and amortizes to 0.078/0.111 ms per step, or 0.026\%/0.018\% of the measured complete step. Added peak allocation is 0.031/0.039 GiB (0.3\% in both settings).

\subsection{Quantized-State AdamW}
\label{sec:method:adamw}

AdamW's second moment $v_t=\operatorname{EMA}(g_t^2)$ is non-negative, strongly heavy-tailed, persistent across steps, and appears in the update denominator. Direct NVFP4 storage therefore distorts small values and destabilizes updates. Hadamard transforms, stochastic rounding, and block/tensor scaling are established FP4 techniques~\citep{quartet2025,hu2025fp4design}; TE likewise applies a tiled $16\times16$ random Hadamard transform to inputs and gradients in its WGRAD recipe~\citep{nvidia_transformer_engine}. We adopt these primitives but do not claim them individually. The optimizer-state pipeline contains no SVD.

For the first moment $m_t$, Hadamard mixing smooths the distribution before SR and adaptive 4/6 quantization. For $v_t$, we use an ordered three-stage preprocessing pipeline before NVFP4 storage:
(1) square root shortens the tail;
(2) a BF16 tile mean separates the positive common component and leaves an approximately zero-mean residual; and
(3) a $16$-element Hadamard transform spreads localized energy and makes the residual more symmetric. Its butterfly structure provides this smoothing at low implementation cost.

The order matters. Mean subtraction without Hadamard leaves most residuals near a negative offset while a few positive values determine the scale. Conversely, applying Hadamard directly to an uncentered non-negative state is likely to create larger transform-domain outliers because same-sign terms, including a common-mode direction, can add constructively. Reconstruction inverts the transform and mean separation and then squares the result.

\textbf{Component reconstruction analysis.}
At one observation point of a 500M run, we quantize and reconstruct extracted FP32 states and average their relative $L_2$ errors. For the first moment, Hadamard reduces the error from 0.0074 to 0.0062 (16.2\%). For the second moment, direct NVFP4 gives 0.0116, whereas the complete pipeline gives 0.0053, a 54.3\% reduction. Removing Hadamard, square root, or tile mean changes the error to 0.0071, 0.0085, and 0.0130, respectively. Thus, holding the other stages fixed, Hadamard reduces error by 25.4\%, while removing the tile mean raises it 12.1\% above direct NVFP4. This single-point analysis measures state reconstruction rather than final training loss. In separate 1B native-state ablations, retaining NVFP4 state iteration while removing any required preprocessing stage caused rapid instability or divergence, so no comparable endpoint loss existed. To our knowledge, this is the first demonstration of stable NVFP4 AdamW momentum-state storage in an integrated 1B pretraining run.

For $N$ elements, the second-moment representation uses $4N$ payload bits, $N$ bits for per-16 BF16 scales, and $N/16$ bits for per-256 BF16 means, totaling $5.0625N$ bits, or \textbf{5.0625 bits/element}. The state is reconstructed rather than consumed by NVFP4 MMA, so BF16 scales improve persistent-state accuracy without constraining Tensor Core input format. Master weights remain BF16; no FP32 master copy is retained.

\begin{figure}[t]
\centering
\begin{subfigure}{0.43\linewidth}
    \centering
    \includegraphics[width=0.93\linewidth]{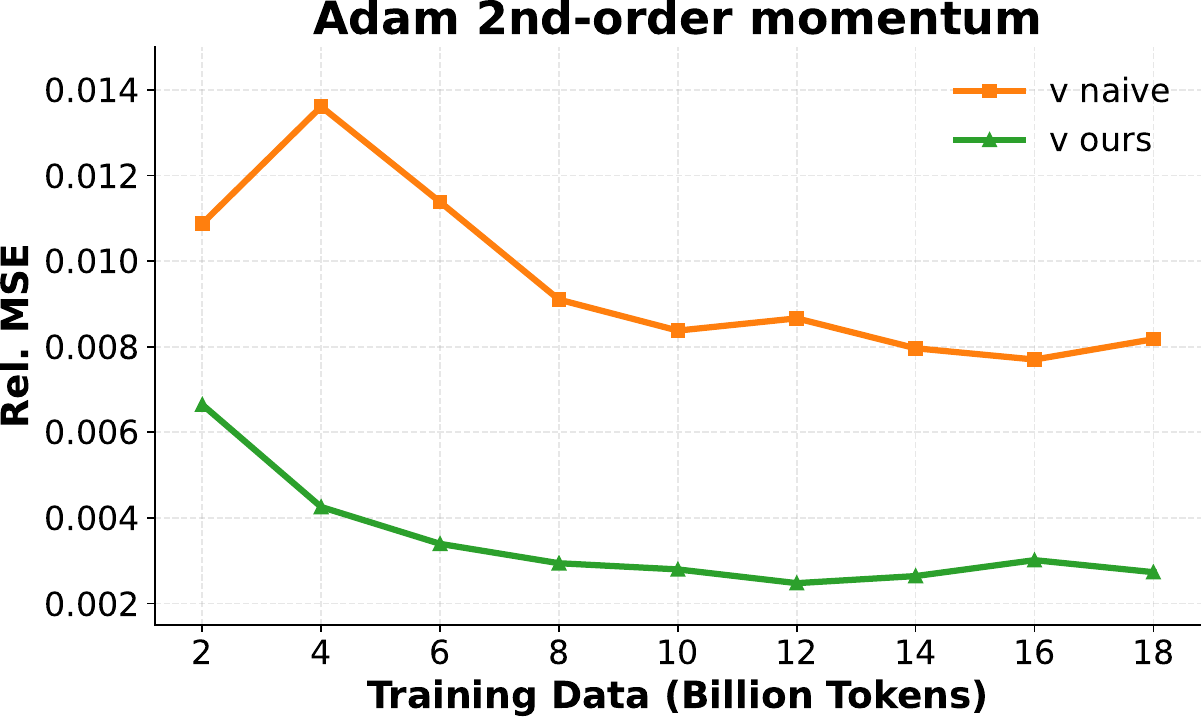}
    \caption{Quantized AdamW}
    \label{fig:adam_mv}
\end{subfigure}
\hfill
\begin{subfigure}{0.43\linewidth}
    \centering
    \includegraphics[width=\linewidth]{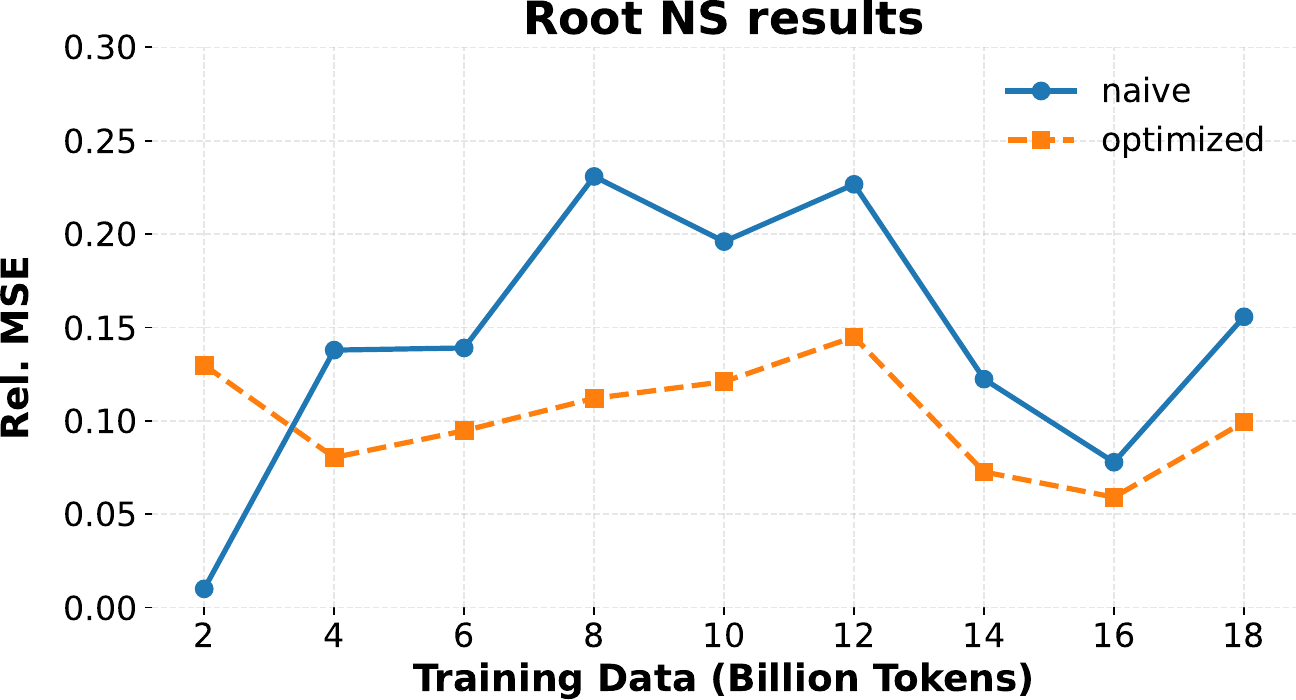}
    \caption{Quantized NS iteration}
    \label{fig:root_ns_coeff}
\end{subfigure}
\caption{\textbf{Optimizer representation error.}
(a) Across the observed 1B training trajectory, the complete transformed-state pipeline reduces second-moment reconstruction error by approximately 60\% relative to direct quantization. The component analysis above separately reports a single-point 500M observation. (b) Shape-optimized coefficients reduce error in NVFP4 Newton--Schulz iterations.}
\label{fig:optimizer_comparison}
\end{figure}

\subsection{NVFP4 Root Optimizer}
\label{sec:method:muon}

Root's Newton--Schulz (NS) update applies repeated matrix products and is sensitive to low-precision iteration error. GRASP/low-bit Muon~\citep{wu2026achieving} stabilizes this computation through a mixed-precision route and a per-step SVD-based subspace operation. We instead combine Root's shape-dependent coefficients with p99 outlier clipping and execute the NS matrix products directly in NVFP4. In our numerical comparison, the relative $L_2$ error decreases from approximately 10\% for a plain all-FP4 route to approximately 3\% for the proposed route.

The first-order momentum follows the first-moment storage recipe in \S\ref{sec:method:adamw}. Q-Root is applied only to eligible matrix parameters. Embeddings, \texttt{lm\_head}, normalization and bias parameters, LoRA factors, and narrow parameters use Q-AdamW. Other hyperparameters follow Root~\citep{root2025}. The NVFP4 NS iteration is
\begin{gather}
  qX_k = Q_{\mathrm{1x16}}(XH_{16}^r),\,
  A_k = qX_k\,{qX_k}^\top 
  \nonumber\\
  qA_k = Q_{\mathrm{1x16}}(A_kH_{16}^r),\,
  qA_k^\mathrm{16x1} = Q_{\mathrm{16x1}}({H_{16}^r}^\top A_k),\, 
  B_k = b A_k + c qA_k qA_k^\mathrm{16x1}
  \nonumber\\
  qB_k = Q_{\mathrm{1x16}}(B_kH_{16}^r),\,
  qX_k^\mathrm{16x1} = Q_{\mathrm{16x1}}({H_{16}^r}^\top X_k),\,
  X_{k+1} = a X_k + qB_k qX_k^\mathrm{16x1}.
  \nonumber
\end{gather}
All quantizations use RTN and adaptive 4/6 scaling. Figure~\ref{fig:root_ns_coeff} shows that the Root coefficients $(a,b,c)$ reduce iteration error relative to standard Muon coefficients. Their aspect-ratio specialization limits error amplification across successive NS steps and avoids an auxiliary mixed-precision or SVD path.

\subsection{Mixed-Precision FP4 Attention}
\label{sec:method:attn}

Existing FP4 attention systems primarily target inference; they do not directly provide a consistent NVFP4 forward--backward operator for pretraining from scratch. In training, $PV$ and $dOV^\top$ are sensitive to low precision, while reused $Q/K/V$ representations must remain consistent between forward and backward.

\begin{figure}[t]
  \centering
  \includegraphics[width=0.85\linewidth]{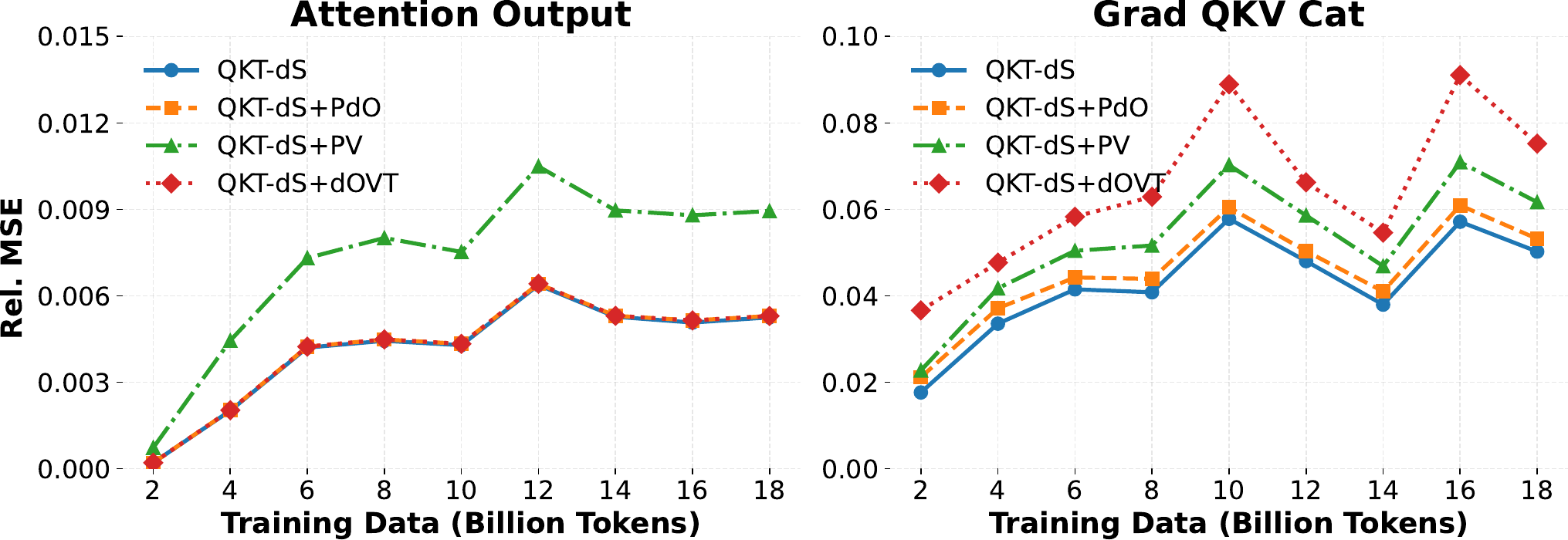}
  \caption{\textbf{Attention precision analysis.}
  Quantizing $PV$ or $dOV^\top$ produces substantially larger error than the selected mixed-precision route, motivating their retention in BF16.}
  \label{fig:attn_precision_split}
\end{figure}

Table~\ref{tab:attn_matmul_precision} assigns NVFP4 to the $QK^\top$ and $dS$ paths and retains the softmax-sensitive $PV$, $P^\top dO$, and $dOV^\top$ products in BF16. Quantized views are reused between forward and backward to avoid precision mismatch and repeated memory traffic. Figure~\ref{fig:attn_precision_split} shows that quantizing $PV$ or $dOV^\top$ degrades accuracy for limited additional savings. Appendix~\ref{app:method_detail:sensitivity_analysis} analyzes these paths, and Appendix~\ref{app:method_detail} gives the tiled algorithm.

\section{Full-Stack NVFP4 Pretraining}
\label{sec:experiments}
\subsection{Experimental Setup}
\label{sec:exp:setup}

We train a 3B Transformer with SwiGLU and GQA on a high-quality subset of Nemotron-CC v2~\citep{nvidia2025nvidianemotronnano2} for 64B tokens. BF16 and Full-Stack FP4 use the same data, 1M-token global batch size, learning-rate schedule, number of steps, and total token budget. The BF16 and fake-quantized runs use two and four A800 80GB nodes, respectively, with eight GPUs per node; the additional nodes compensate for the slower fake-quantization simulation and do not change the global optimization settings. Every Transformer block is quantized in the FP4 run. Appendix~\ref{app:model-data} gives the architecture and hyperparameters.

Native-hardware efficiency is evaluated separately on one RTX 5090 (SM120) using four target-shaped decoder blocks. These measurements are not the throughput or memory of the complete distributed 3B run.

\subsection{Ablations and Modular Adoption}
\label{sec:exp:ablation}

We evaluate module-level ablations on a 1B model trained for 20B tokens. NVFP4+4/6~\citep{fouroversix2025} is the matched linear-only quality baseline. All quantized rows apply the selected recipes to every block rather than reserving BF16 blocks. ``Full-Stack w/o Root'' combines LoRA-SVD, Q-Attn, and Q-AdamW; in the Root rows, Q-Root handles eligible matrices and Q-AdamW handles the remaining parameters.

\begin{table}[t]
\begin{minipage}[t]{0.43\linewidth}
  \vspace{0pt}
  \centering
  \captionof{table}{\textbf{Precision assignment in Q-Attn.} Hats denote quantized views reused across passes.}
  \label{tab:attn_matmul_precision}
  \vspace{5pt}
  \scriptsize
  \begin{tabular}{lcc}
    \toprule
    \textbf{Matmul} & \textbf{Pass} & \textbf{Precision} \\
    \midrule
    $\hat{Q}\hat{K}^\top$    & Fwd & NVFP4 \\
    $P\hat{V}$               & Fwd & BF16 \\
    $\hat{Q}\hat{K}^\top$    & Bwd & NVFP4 \\
    $P^\top dO$              & Bwd & BF16 \\
    $dO \,\hat{V}^\top$      & Bwd & BF16 \\
    $\hat{dS}\,\hat{K}$      & Bwd & NVFP4 \\
    $\hat{dS}^\top\hat{Q}$   & Bwd & NVFP4 \\
    \bottomrule
  \end{tabular}
\end{minipage}\hspace{0.02\linewidth}
\begin{minipage}[t]{0.52\linewidth}
  \vspace{0pt}
  \centering
  \captionof{table}{\textbf{Ablations on 1B/20B-token pretraining.} All quantized recipes are applied to every block.}
  \label{tab:ablation}
  \vspace{3pt}
  \scriptsize
  \begin{tabular}{lcc}
      \toprule
      \textbf{Configuration} & \textbf{Loss $\downarrow$} & \textbf{$\Delta$BF16 $\downarrow$} \\
      \midrule
      \multicolumn{3}{l}{\emph{AdamW only}} \\
      BF16 & 2.6415 & --- \\
      BF16 + LoRA & \textbf{2.6351} & \textbf{$-0.24\%$} \\
      NVFP4 4/6 & 2.6785 & +1.40\% \\
      4/6 + plain LoRA & 2.6711 & +1.12\% \\
      4/6 + LoRA-SVD & \textbf{2.6575} & \textbf{+0.61\%} \\
      4/6 + Q-AdamW & 2.6674 & +0.98\% \\
      4/6 + Q-Attn & 2.6539 & +0.47\% \\
      \textbf{Full-Stack w/o Root} & \textbf{2.6563} & \textbf{+0.56\%} \\
      \midrule
      \multicolumn{3}{l}{\emph{Root + AdamW}} \\
      BF16 + Root & 2.4819 & --- \\
      \textbf{Full-Stack} & \textbf{2.4973} & \textbf{+0.62\%} \\
      \bottomrule
    \end{tabular}
\end{minipage}
\end{table}

Plain LoRA reduces the linear-only gap from 1.40\% to 1.12\%, whereas LoRA-SVD reduces it to 0.61\%. The plain-LoRA change therefore closes 0.28 of the observed 0.79 percentage-point gap reduction (35.4\%); adding SVD initialization and subspace-constrained routing closes a further 0.51 points (64.6\%). In the no-quantization control, BF16+LoRA reduces loss from 2.6415 to 2.6351 ($-0.24\%$). These controls show a positive but comparatively small capacity contribution under the evaluated settings.

LoRA-SVD increases parameters by 4.32\% at 1B, from 1,212,663,808 to 1,265,092,608, and by 3.12\% at 3B. Both model sizes use rank 128 only for \texttt{down\_proj} and rank 64 for \texttt{gate\_up\_proj}, concatenated \texttt{qkv\_proj}, and \texttt{out\_proj}; rank does not scale with model size. The progressive rows in Table~\ref{tab:ablation} also show that the recipes can be enabled independently and composed in one training run.

\subsection{3B Pretraining and Downstream Evaluation}
\label{sec:exp:main}

\begin{figure}[t]
  \centering
  \includegraphics[width=0.85\linewidth]{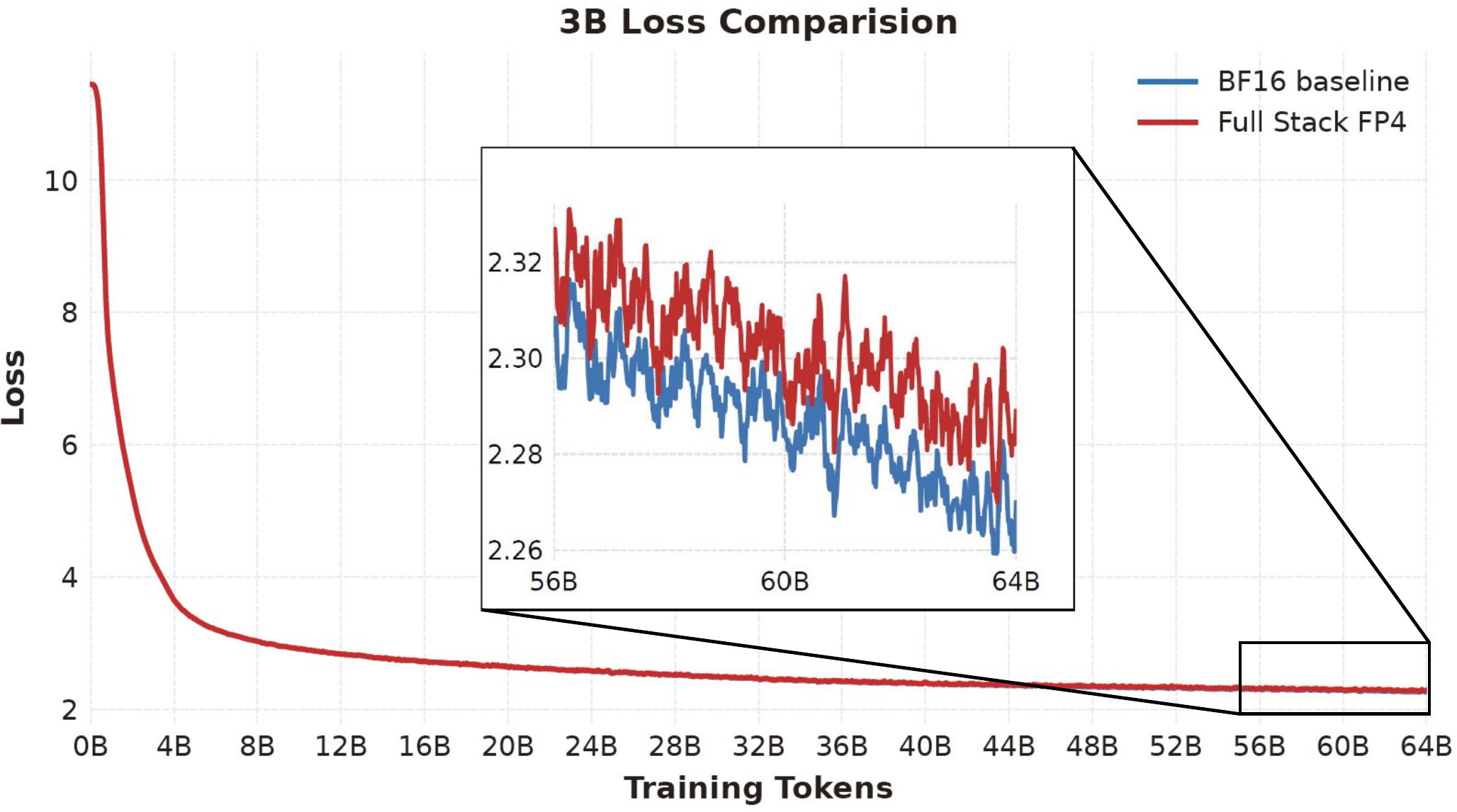}
  \caption{\textbf{Training loss on 3B/64B-token pretraining.}
  Under matched data and optimization budgets, BF16 and Full-Stack FP4 reach losses of 2.267 and 2.286, respectively, a 0.838\% gap.}
  \label{fig:main_training_curves}
\end{figure}

Figure~\ref{fig:main_training_curves} reports the matched 3B runs. The BF16 Root+AdamW baseline reaches 2.267 and Full-Stack FP4 reaches 2.286, corresponding to a \textbf{0.838\%} relative gap. The two loss trajectories remain close over the evaluated 64B-token training horizon.

We further evaluate the final checkpoints zero-shot with \texttt{max\_length=2048}. ARC-C and HellaSwag use normalized accuracy. Table~\ref{tab:downstream_3b} shows that Full-Stack FP4 has slightly lower average perplexity (26.665 versus 26.675) and 0.10 percentage points lower average accuracy. The individual accuracy differences range from 0.03 to 0.16 percentage points.

\begin{table}[t]
\centering
\caption{\textbf{Zero-shot evaluation of the 3B checkpoints.} The vertical rule separates perplexity and accuracy metrics.}
\label{tab:downstream_3b}
\small
\resizebox{\linewidth}{!}{%
\begin{tabular}{lccc|cccccc}
\toprule
\textbf{Model} & \textbf{Wiki} & \textbf{LAMBADA} & \textbf{Avg. PPL} & \textbf{ARC-E} & \textbf{ARC-C} & \textbf{HellaSwag} & \textbf{LAMBADA} & \textbf{PIQA} & \textbf{Avg. Acc.} \\
& \textbf{PPL $\downarrow$} & \textbf{PPL $\downarrow$} & $\downarrow$ & $\uparrow$ & \textbf{Norm. $\uparrow$} & \textbf{Norm. $\uparrow$} & $\uparrow$ & $\uparrow$ & $\uparrow$ \\
\midrule
BF16 & 26.24 & \textbf{27.11} & 26.675 & \textbf{.6272} & \textbf{.3020} & \textbf{.4695} & \textbf{.3929} & \textbf{.7016} & \textbf{.49864} \\
Full-Stack FP4 & \textbf{26.09} & 27.24 & \textbf{26.665} & .6259 & .3017 & .4686 & .3913 & .7007 & .49764 \\
\bottomrule
\end{tabular}}
\end{table}

\subsection{Native-Hardware Efficiency}
\label{sec:exp:efficiency}

We measure four decoder blocks on one RTX 5090 (SM120) with $B=2$, $L=8192$ (16,384 tokens), excluding embeddings and the LM head. The $d=4096/6144$ shapes use $d_{ff}=11008/13824$, 32/48 query heads, and 8 KV heads, representing approximately 0.7B/1.4B dense block parameters. Each model row uses five warmups and 20 measurements in a fresh process; isolated optimizer rows use 20 warmups and 100 measurements. CUDA-event segments are synchronized. Accounting includes LoRA gradients in forward/backward, state transformations, packed AdamW updates, QR refresh, five Root NS iterations, and parameter updates; periodic fast-SVD is reported separately. SM120 lacks an efficient stochastic-rounding primitive on this path, so TE and ours both disable SR.

BF16 is a strong full-graph, max-autotune \texttt{torch.compile} baseline with CODA-style epilogue fusion~\citep{guo2026coda}, not eager PyTorch. TE 2.17 is NVIDIA's official fused NVFP4 path with RHT and no LoRA~\citep{nvfp4,nvidia_transformer_engine}. Ours includes fused NVFP4, RHT, adaptive 4/6, and the complete LoRA-SVD path. Four Over Six remains the projection-quality baseline, but its public training implementation lacks complete fused-CUDA coverage; on the evaluated shapes, its less-fused Triton backward path was slower than eager BF16. We therefore use TE as the optimized systems baseline.

\begin{table}[t]
\centering
\caption{\textbf{Complete four-block local training step.} Each cell reports milliseconds / peak allocated GiB. Forward, backward, and optimizer are segments of the same step.}
\label{tab:native_complete_step}
\small
\begin{tabular}{lccc}
\toprule
\textbf{Optimizer/shape} & \textbf{BF16} & \textbf{TE} & \textbf{Ours} \\
\midrule
Root, $d=4096$ & 760.16 / 12.35 & 556.25 / 12.75 & \textbf{303.84 / 9.70} \\
Root, $d=6144$ & 1750.27 / 18.55 & 1343.33 / 19.11 & \textbf{618.87 / 14.27} \\
AdamW, $d=4096$ & 406.26 / 16.31 & \textbf{199.61} / 16.71 & 200.66 / \textbf{10.13} \\
AdamW, $d=6144$ & 761.31 / 26.22 & 330.07 / 26.77 & \textbf{326.36 / 15.08} \\
\bottomrule
\end{tabular}
\end{table}

In the Root setting, BF16 and TE apply Muon to dense matrices, while ours applies Q-Root to base matrices and packed Q-AdamW to LoRA factors. Ours is 2.50$\times$/2.83$\times$ faster than BF16 and 1.83$\times$/2.17$\times$ faster than TE, with 21.5\%/23.1\% less peak memory than BF16. In the AdamW setting, ours is 0.5\% slower/1.1\% faster than TE and uses 37.9\%/42.5\% less memory than BF16. TE retains high-precision resident and optimizer states while adding NVFP4 views, scales, RHT workspace, and saved tensors; ours compresses the optimizer states and reuses workspaces despite including LoRA.

At the same single-GPU batch size and input shape, the Root step ratios correspond to 60.0\%/64.6\% less time for the same number of tokens than BF16. An approximate interpolation of the 3B loss--token curves indicates 7.8\% additional FP4 tokens to reach the BF16 loss. Under the same fixed-input assumption, combining this factor with the local timings gives estimated matched-loss time savings of 56.9\%/61.9\%. These are cross-setting estimates, not measured complete-model or distributed loss-matched runs. They also do not account for increasing batch size or throughput by filling the memory freed by FP4; we therefore do not report a fixed-memory or capacity-matched efficiency result. Appendix~\ref{app:native_efficiency} reports component measurements and current kernel limitations.

\section{Related Work}
\label{sec:related}
\paragraph{FP8 and FP4 pretraining.}
Early low-bit training includes WAGE~\citep{wu2018training}, FP8 training with E4M3/E5M2~\citep{micikevicius2022fp8}, and stable 8-bit vision--language training~\citep{wortsman2023stable}. FP8 systems later expanded coverage to activations and optimizer states~\citep{fishman2024scaling,coat2024}, and DeepSeek-V3 demonstrated large-scale FP8 pretraining~\citep{deepseekv3}. For 4-bit training, MXFP introduced the MXFP4 format~\citep{mxfp2023}, while NVIDIA's NVFP4 recipe combined fine-grained scaling, stochastic rounding, and Hadamard transforms~\citep{nvfp4}. Transformer Engine provides NVIDIA's optimized implementation of this projection path~\citep{nvidia_transformer_engine}.

Subsequent work improves projection quality or training behavior. FP4-All-the-Way~\citep{fp4alltheway2025} studies fully quantized projection training and the decreasing gradient-signal-to-quantization-noise ratio over training. Quartet~\citep{quartet2025} and the FP4 design-space study~\citep{hu2025fp4design} analyze scaling, Hadamard transforms, stochastic rounding, and implementation trade-offs. Four Over Six~\citep{fouroversix2025} introduces adaptive $M=4/M=6$ block scaling and is our matched projection-quality baseline. TetraJet and TetraJet-v2~\citep{tetrajet2025,chen2025tetrajetv2}, CHON~\citep{chon2026}, and Metis~\citep{cao2025metis} address weight oscillation, outliers, or anisotropic distributions. These methods primarily improve projection-side quantization. Our work instead extends NVFP4 recipes to persistent optimizer states, optimizer computation, and a trainable attention path while retaining Four Over Six for projection quality.

\paragraph{Low-bit optimizers.}
Block-wise INT8 Adam~\citep{dettmers2022int8}, FP8 optimizer states in COAT~\citep{coat2024}, and 4-bit Adam with rank-1 normalization~\citep{li2024memory4bit} reduce optimizer-state memory. Quantized Muon has been studied with INT8 momentum~\citep{gupta2025muonquant} and with GRASP's grid-quantized mixed-precision stabilization and SVD-based subspace operation~\citep{wu2026achieving}. Our Q-AdamW contribution is an NVFP4 representation for both persistent momentum states that remains stable in integrated pretraining. Q-Root is a separate direct NVFP4 Newton--Schulz route based on Root's shape-dependent coefficients and clipping.

\paragraph{Low-bit attention.}
The SageAttention series~\citep{zhang2024sageattention,zhang2024sageattention2,sageattn3_2025} develops INT8 and FP4 attention kernels mainly for inference; SageAttention3's training path uses INT8 rather than NVFP4. Attn-QAT~\citep{attnqat2026} trains models for quantized inference and identifies forward--backward recomputation consistency, but it does not provide an NVFP4 attention operator for pretraining from scratch. These methods build on the IO-aware tiling of FlashAttention~\citep{dao2022flashattention,dao2023flashattention2}; FlashAttention-4~\citep{zadouri2026flashattention4} further introduces Blackwell-oriented pipeline optimizations. Our Q-Attn instead studies a trainable NVFP4/BF16 precision split and evaluates its forward, backward, memory, and convergence behavior.

\paragraph{Systems baselines.}
CODA rewrites Transformer blocks as fused GEMM--epilogue programs~\citep{guo2026coda}; we use the same systems principle with full-graph, max-autotune \texttt{torch.compile} for the BF16 baseline. NVIDIA TE 2.17 is the primary systems baseline because it provides NVIDIA's fused NVFP4 implementation. Four Over Six is the stronger projection-quality baseline, but its current public end-to-end training path has less complete CUDA fusion, especially in backward. We therefore use TE for systems measurements and Four Over Six for projection-quality comparisons.

\section{Conclusion and Limitations}
\label{sec:conclusion}

Full-Stack FP4 provides independent NVFP4 recipes for projections, optimizer states, Root computation, and attention. Their combined 3B/64B-token run remains within \textbf{0.838\%} of the BF16 training loss and yields similar aggregate zero-shot metrics. Native four-block measurements further show substantial Root speedups and AdamW memory savings on one RTX 5090. The evidence is limited to models up to 3B parameters and does not establish longer-horizon or larger-scale behavior.

\textbf{Outlook on module- and stage-wise precision.}
Because gradient signals can weaken relative to quantization noise over training~\citep{fp4alltheway2025}, LoRA-SVD's error reduction may extend the useful NVFP4 phase; validating this hypothesis requires longer runs. A practical system could enable low precision by module and stage, restore selected paths to higher precision when needed, and merge LoRA-SVD into dense weights or retain it as an adapter. Such a design could assign precision independently to projections, attention, gradients, and optimizer components according to the training stage and available hardware.

\textbf{Limitations.}
\label{sec:limitations}
Quality experiments cover at most 3B parameters and 64B tokens; they do not establish behavior at 7B+ scale or over longer training. Native efficiency is measured on four target-shaped blocks on one RTX 5090, not on a complete or distributed model. The matched-loss time calculation is a cross-setting estimate at fixed single-GPU batch size and does not account for increasing batch size by using the memory freed by FP4. LoRA backward and attention kernels remain incompletely fused, and the attention implementation does not yet use several FlashAttention-4 optimizations. Finally, the present recipes and measurements are specific to NVFP4 and do not establish transfer to MXFP4.

\FloatBarrier
\bibliographystyle{plainnat_etal4}
\bibliography{refs}

\FloatBarrier
\newpage
\appendix

\section{Full Derivation of LoRA-SVD}
\label{app:method_linear_detail}

The aligned forward and naive backward computations are
\begin{gather}
Y = \underbrace{Q_\mathrm{1x16}(X)\,Q_\mathrm{16x16}(W_\mathrm{res})^\top}_{\text{NVFP4 GEMM}}
    + \,\beta \underbrace{(XL_1^\top)L_2^\top}_{\text{BF16}}
  \nonumber\\
  \mathrm{d} X = Q_{\mathrm{1x16}}(\mathrm{d} Y)\,Q_{\mathrm{16x16}}(W_{\mathrm{res}}) + \beta (\mathrm{d}Y L_2)L_1
  \nonumber\\
  \mathrm{d} W = Q_{\mathrm{1x16}}(\mathrm{d} Y^\top H_{16}^r) \, Q_{\mathrm{16x1}}\left({H_{16}^r}^\top Q_{\mathrm{1x16}}(X)\right)
  \nonumber\\
  \mathrm{d} W_{\mathrm{res}} = \mathrm{d} W
  \nonumber\\
  \mathrm{d} L_1 = \beta L_2^\top (\mathrm{d} Y^\top X),\qquad
  \mathrm{d} L_2 = \beta (\mathrm{d} Y^\top X) L_1^\top .
  \nonumber
\end{gather}

The forward pass uses RTN and the backward pass uses SR; both use adaptive 4/6 scaling. $H_{16}^r$ is a fresh $16\times16$ random Hadamard transform. Computing $\mathrm{d}L_1$ and $\mathrm{d}L_2$ naively repeats the large GEMM used for $\mathrm{d}W_{\mathrm{res}}$. We instead reuse the already formed weight gradient:
\begin{equation}
  \mathrm{d} L_1 = \beta L_2^\top \mathrm{d} W ,\qquad
  \mathrm{d} L_2 = \beta \mathrm{d} W L_1^\top .
  \nonumber
\end{equation}
This approximation avoids another large GEMM over full-precision activations. The following subspace constraint separates the components assigned to the two branches.

Cholesky-QR gives
\begin{equation}
  Q_{L_2} R_{L_2} = L_2, \qquad Q_{L_1} R_{L_1} = L_1^\top,
  \nonumber
\end{equation}
where $R=\operatorname{chol}(A^\top A)^\top\in\mathbb{R}^{r\times r}$ and $Q=AR^{-1}\in\mathbb{R}^{d\times r}$. For $r\ll d$, this operates on small $r\times r$ matrices. We then compute
\begin{gather}
  \Lambda = Q_{L_2}^\top\mathrm{d}WQ_{L_1}\in\mathbb{R}^{r\times r}
  \nonumber\\
  \mathrm{d}W_{\mathrm{LoRA}} = Q_{L_2}\Lambda Q_{L_1}^\top
  \nonumber\\
  \mathrm{d}W_{\mathrm{res}} = \mathrm{d}W-\mathrm{d}W_{\mathrm{LoRA}}
  \nonumber\\
  \mathrm{d}L_1 = \beta L_2^\top\mathrm{d}W_{\mathrm{LoRA}},\qquad
  \mathrm{d}L_2 = \beta\mathrm{d}W_{\mathrm{LoRA}}L_1^\top .
  \nonumber
\end{gather}

\section{Kernel Fusion of LoRA-SVD}
\label{app:method_linear_kernel}

The $Y$ and $\mathrm{d}X$ paths read high-precision $X$ or $\mathrm{d}Y$ and write large outputs. Their low-rank operations can therefore share input and output traffic with adjacent quantization and NVFP4 GEMM kernels. The activation quantizer can be fused with normalization or SwiGLU, while output projections still require a dedicated quantization path. Similar reuse applies to $\mathrm{d}Y$.

SVDQuant~\citep{li2024svdquant}, originally developed for post-training quantization and inference, shows that a separate low-rank branch can add latency through repeated movement of 16-bit activations. Its Nunchaku engine fuses low-rank computation with quantization and 4-bit kernels to remove intermediate memory traffic and redundant launches. LoRA-SVD has the same opportunity in $Y$ and $\mathrm{d}X$, but the current training backward is not fully fused. In Appendix~\ref{app:native_efficiency}, the LoRA-SVD MLP is 11.2--17.2\% slower than TE while remaining 2.87--3.30$\times$ faster than BF16. An operator model projects below-5\% overhead after full fusion; this is an estimate rather than a measured result.

\section{Quantization Sensitivity in Attention}
\label{app:method_detail:sensitivity_analysis}

This section explains the empirical sensitivity of $PV$ and $dO V^\top$ in Figure~\ref{fig:attn_precision_split} through error terms induced by block quantization. Their magnitude depends on the attention distribution.

\subsection{Sensitivity of $PV$}

The attention matrix $P\in\mathbb{R}^{N\times N}$ satisfies $\sum_jP_{i,j}=1$ and $P_{i,j}\geq0$. In many heads, a small number of entries carry much of the probability mass while other entries are small. Under block quantization, a local maximum sets the scale, so smaller probabilities in the same block may be represented coarsely or rounded to zero.

Without explicit renormalization, this can change the row sum and perturb the convex combination. Writing $\Delta P=\hat P-P$, the output error is
\begin{equation}
\Delta O_i=\Delta P_iV,\qquad
\|\Delta O_i\|\leq\|\Delta P_i\|_1\max_j\|V_j\|.
\nonumber
\end{equation}
The skewed distribution therefore makes $PV$ sensitive in the evaluated setting, consistent with Figure~\ref{fig:attn_precision_split}.

\subsection{Sensitivity of $dOV^\top$}

In a FlashAttention-style backward pass,
\begin{equation}
    dS_{i,j}=P_{i,j}(dP_{i,j}-D_i),\qquad
    dP_{i,j}=\langle dO_i,V_j\rangle,
    \nonumber
\end{equation}
where $D_i=\sum_kP_{i,k}dP_{i,k}=\langle O_i,dO_i\rangle$. If quantizing $dOV^\top$ introduces error $\Delta dP_{i,j}$ while $D_i$ remains in higher precision, then
\begin{equation}
    \label{eq:spurious_grad}
    \widehat{dS}_{i,j}=dS_{i,j}+P_{i,j}\Delta dP_{i,j}.
\end{equation}

Block scaling is set by the largest magnitude in each group, so smaller $dP$ entries can receive larger relative error. The ordering of $|dP|$ need not match that of $P$. For an index $c$ with appreciable probability but a sub-maximum $dP_{i,c}$,
\begin{equation}
    \widehat{dS}_{i,c}=dS_{i,c}+P_{i,c}\Delta dP_{i,c}.
    \nonumber
\end{equation}
The perturbation can be large relative to $dS_{i,c}$ when $dP_{i,c}-D_i$ is small. Conversely, the factor $P$ attenuates this term in the dense tail. Retaining $dOV^\top$ in BF16 avoids this error source in our evaluated precision split.

\section{Mixed-Precision Attention Algorithm}
\label{app:method_detail}

Algorithm~\ref{alg:mixattn} summarizes the tiled forward and backward pass. $Q$, $K$, and $V$ are stored as quantized views; sensitive products reconstruct their inputs and execute in BF16.

\begin{algorithm}[t]
\caption{Mixed-Precision NVFP4 Attention (one head)}
\label{alg:mixattn}
\footnotesize
\begin{algorithmic}[1]
\STATE {\bfseries Require} $Q,K,V\in\mathbb{R}^{N\times d}$ (BF16), tile sizes $B_q,B_k$
\STATE {\bfseries Require} NVFP4 quantizer $\phi(\cdot)$
\STATE \textbf{// Forward pass}
\STATE $\hat Q\gets\phi(Q)$; $\hat K\gets\phi(K)$; $\hat V\gets\phi(V)$
  \hfill\COMMENT{NVFP4 views}
\STATE Save $\hat Q,\hat K,\hat V$ to HBM
  \hfill\COMMENT{compact NVFP4 views with scales}
\FOR{$i=1$ to $T_q$}
  \STATE $m_i\gets-\infty$; $l_i\gets0$; $O_i\gets0$
  \FOR{$j=1$ to $T_k$}
    \STATE $S_{ij}\gets\hat Q_i\hat K_j^\top/\sqrt d$
      \hfill\COMMENT{NVFP4 GEMM, FP32 accumulation}
    \STATE Apply causal mask
    \STATE $\tilde P_{ij}\gets\operatorname{softmax}_{\mathrm{online}}(S_{ij})$
    \STATE $O_i\gets\operatorname{diag}(\alpha)O_i+\tilde P_{ij}\hat V_j$
      \hfill\COMMENT{dequantized $\hat V$, BF16 GEMM}
  \ENDFOR
  \STATE $L_i\gets m_i+\log l_i$
\ENDFOR
\STATE \textbf{// Backward pass} (given $dO$ in BF16)
\STATE $D\gets\operatorname{rowsum}(dO\odot O)$
\FOR{$j=1$ to $T_k$}
  \FOR{$i=1$ to $T_q$}
    \STATE Recompute $S_{ij}\gets\hat Q_i\hat K_j^\top/\sqrt d$
    \STATE Recompute $P_{ij}$ from $S_{ij}$ and $L_i$
    \STATE $dV_j\mathrel{+}=P_{ij}^\top dO_i$ \hfill\COMMENT{BF16}
    \STATE $dS_{ij}\gets P_{ij}\odot(dO_i\hat V_j^\top-D_i)$
      \hfill\COMMENT{dequantized $\hat V$, BF16 GEMM}
    \STATE $\widehat{dS}_{ij}\gets\phi_{\mathrm{SR}}(dS_{ij})$
    \STATE $dQ_i\mathrel{+}=\widehat{dS}_{ij}\hat K_j/\sqrt d$
      \hfill\COMMENT{NVFP4 GEMM}
    \STATE $dK_j\mathrel{+}=\widehat{dS}_{ij}^\top\hat Q_i/\sqrt d$
      \hfill\COMMENT{NVFP4 GEMM}
  \ENDFOR
\ENDFOR
\STATE {\bfseries Return} $O,dQ,dK,dV$
\end{algorithmic}
\end{algorithm}

\paragraph{Distinction from Attn-QAT.}
Attn-QAT~\citep{attnqat2026} uses QAT to obtain quantized attention for inference rather than an NVFP4 forward--backward pretraining operator. Our design targets pretraining: it stores quantized $Q$, $K$, and $V$ views, uses NVFP4 for $QK^\top$ and the $dS$ products, and executes $PV$ and $dOV^\top$ in BF16.

\section{Native-Hardware Efficiency Details}
\label{app:native_efficiency}

Table~\ref{tab:native_components} decomposes the complete-step measurements in Table~\ref{tab:native_complete_step}. Component memory is incremental peak allocation rather than complete training memory. Both NVFP4 MLP implementations create quantized views, scales, RHT workspace, and saved tensors absent in BF16. Ours uses compact quantized intermediates and releases temporary buffers earlier than TE.

\begin{table}[H]
\centering
\caption{\textbf{Native component measurements on one RTX 5090.} Each cell reports milliseconds / incremental peak allocated GiB.}
\label{tab:native_components}
\small
\begin{tabular}{lccc}
\toprule
\textbf{Region/shape} & \textbf{BF16} & \textbf{Comparator} & \textbf{Ours} \\
\midrule
MLP, $d=4096$ & 61.79 / 2.14 & TE: \textbf{18.38} / 3.35 & 21.54 / \textbf{2.63} \\
MLP, $d=6144$ & 115.92 / 2.83 & TE: \textbf{31.60} / 4.44 & 35.13 / \textbf{3.24} \\
Attention, $d=4096$ & 18.87 / 1.379 & --- & \textbf{14.90 / 0.675} \\
Attention, $d=6144$ & 28.52 / 2.006 & --- & \textbf{22.49 / 0.945} \\
Root, $d=4096$ & 344.55 / 4.42 & GRASP: 474.16 / 6.21 & \textbf{109.80 / 3.25} \\
Root, $d=6144$ & 981.64 / 8.54 & GRASP: 1239.20 / 11.85 & \textbf{304.28 / 6.35} \\
AdamW, $d=4096$ & 10.21 / 7.92 & --- & \textbf{7.28 / 3.48} \\
AdamW, $d=6144$ & 19.73 / 15.33 & --- & \textbf{14.04 / 6.73} \\
\bottomrule
\end{tabular}
\end{table}

\paragraph{LoRA-SVD MLP.}
The MLP rows exclude attention and optimizer computation but include LoRA gradients. Ours is 17.2\%/11.2\% slower than TE and 2.87$\times$/3.30$\times$ faster than BF16. Adaptive 4/6 adds approximately 0.01\%; the remaining TE gap is primarily the not-yet-fully-fused LoRA backward path. SVDQuant~\citep{li2024svdquant} demonstrates that fusing low-rank computation with quantization and low-bit kernels removes redundant traffic. Our operator model projects below-5\% overhead after full fusion; this is projected headroom, not a measured result.

\paragraph{Attention.}
The attention rows compare BF16 SDPA with the NVFP4 Q/K body; QKV and output projections are counted in the complete-step table. Ours is 1.27$\times$ faster and uses 51.1\%/52.9\% less incremental allocation. The training kernels remain under optimization: LoRA backward is not yet fully fused, and the attention path does not yet incorporate several FlashAttention-4 pipeline optimizations~\citep{zadouri2026flashattention4}.

\paragraph{Root.}
Root rows match shapes, p99 clipping, aspect-ratio coefficients, and five Newton--Schulz iterations. GRASP uses its published method and official implementation~\citep{wu2026achieving}. Ours is 3.14$\times$/3.23$\times$ faster than BF16 and 4.32$\times$/4.07$\times$ faster than GRASP, with 26.4\%/25.6\% less allocation than BF16.

\paragraph{AdamW.}
The isolated rows use precomputed synchronized gradients and exclude LoRA, QR, and projection computation. Both implementations fuse decay and parameter update. BF16 stores a BF16 weight and gradient and two FP32 moments, totaling 96 bits/parameter. Ours uses a 5-bit first moment and a 5.0625-bit second moment with the same BF16 weight and gradient, totaling 42.0625 bits/parameter, with no FP32 master weight. AdamW updates are memory-bandwidth bound, so reduced moment traffic outweighs packing and reconstruction: ours is 1.40$\times$/1.41$\times$ faster with 56.0\%/56.1\% less allocation.

\paragraph{Periodic fast-SVD.}
With a 2,048-step period, one realignment costs 160.00/226.96 ms at $d=4096/6144$. This amortizes to 0.078/0.111 ms per step, or 0.026\%/0.018\% of the complete ours step. Added peak allocation is 0.031/0.039 GiB (0.3\% in both settings).

\paragraph{Measurement scope.}
These results measure one GPU and four target-shaped blocks, not complete-model or distributed throughput. The same-token and matched-loss estimates in \S\ref{sec:exp:efficiency} keep the single-GPU batch size fixed and do not exploit the additional batch capacity made possible by lower FP4 memory use.

\section{Model Architecture and Training Configuration}
\label{app:model-data}

Tables~\ref{tab:model_arch} and~\ref{tab:training_hparams} summarize the 3B architecture and matched training settings.

\begin{table}[H]
\centering
\begin{minipage}{0.48\textwidth}
  \centering
  \small
  \begin{tabular}{lc}
    \toprule
    \textbf{Model architecture} & \textbf{Value} \\
    \midrule
    Number of layers & 24 \\
    Hidden dimension & 3072 \\
    FFN intermediate dimension & 9216 \\
    Attention heads & 24 \\
    Attention KV heads & 12 \\
    Context window & 2048 tokens \\
    \bottomrule
  \end{tabular}
  \caption{3B model architecture.}
  \label{tab:model_arch}
\end{minipage}
\hfill
\begin{minipage}{0.48\textwidth}
  \centering
  \small
  \begin{tabular}{lc}
    \toprule
    \textbf{Training configuration} & \textbf{Value} \\
    \midrule
    Optimizer & Root+AdamW \\
    Peak learning rate & $1.5 \times 10^{-4}$ \\
    LoRA peak learning rate & $3.0 \times 10^{-4}$ \\
    Weight decay & 0.1 \\
    $\beta_1 / \beta_2$ & 0.9 / 0.95 \\
    Gradient clipping & 1.0 \\
    Warmup tokens & 3B tokens (5\%) \\
    Learning-rate decay & Cosine \\
    Global batch size & 1M tokens \\
    Root quantile ratio & 0.99 \\
    Training tokens & 64B \\
    \bottomrule
  \end{tabular}
  \caption{Matched training hyperparameters.}
  \label{tab:training_hparams}
\end{minipage}
\end{table}

\begin{table}[H]
\centering
\caption{\textbf{LoRA-SVD ranks and 3B parameter increases.} The same rank policy is used at 1B and 3B.}
\label{tab:lora_rank_policy}
\small
\begin{tabular}{lcc}
\toprule
\textbf{Projection} & \textbf{Rank} & \textbf{3B parameter increase} \\
\midrule
\texttt{down\_proj} & 128 & 5.56\% \\
\texttt{gate\_up\_proj} & 64 & 2.43\% \\
Concatenated \texttt{qkv\_proj} & 64 & 3.13\% \\
\texttt{out\_proj} & 64 & 4.17\% \\
\midrule
Overall & --- & 3.12\% \\
\bottomrule
\end{tabular}
\end{table}

\end{document}